%% file: paper.tex
\documentclass[10pt,twocolumn,letterpaper]{article}
\usepackage[pagenumbers]{wacv}
\usepackage{graphicx}
\usepackage{amsmath}
\usepackage{amssymb}
\usepackage{booktabs}
\usepackage{comment}
\usepackage{color}
\usepackage{enumitem}
\usepackage{multirow}
\usepackage[pagebackref,breaklinks,colorlinks]{hyperref}
\usepackage[capitalize]{cleveref}
\providecommand{\keywords}
{
  \small	
  \textbf{\textit{Keywords---}} 
}

\begin{document}
\title{DepGAN : Leveraging Depth Maps for Handling Occlusions and Transparency in Image Composition}

\author{Amr Ghoneim\\
Saint Mary's University, Halifax, Canada\\
{\tt\small amrtsg@gmail.com}
\and
Jiju Poovvancheri\\
Saint Mary's University, Halifax, Canada\\
{\tt\small jiju.poovvancheri@smu.ca}
\and
Yasushi Akiyama\\
Saint Mary's University, Halifax, Canada\\
{\tt\small yasushi.akiyama@smu.ca}
\and
Dong Chen\\
Nanjing Forestry University, Nanjing, China.\\
{\tt\small chendong@njfu.edu.cn}
}
\maketitle

\begin{abstract}
Image composition is a complex task which requires a lot of information about the scene for an accurate and realistic composition, such as perspective, lighting, shadows, occlusions, and object interactions. Previous methods have predominantly used 2D information for image composition, neglecting the potentials of 3D spatial information. In this work, we propose DepGAN, a Generative Adversarial Network that utilizes depth maps and alpha channels to rectify inaccurate occlusions and enhance transparency effects in image composition. Central to our network is a novel loss function called Depth Aware Loss which quantifies the pixel wise depth difference to accurately delineate occlusion boundaries while compositing objects at different depth levels. Furthermore, we enhance our network's learning process by utilizing opacity data, enabling it to effectively manage compositions involving transparent and semi-transparent objects. We tested our model against state-of-the-art image composition GANs on benchmark (both real and synthetic) datasets. The results reveal that DepGAN significantly outperforms existing methods in terms of accuracy of object placement semantics, transparency and occlusion handling, both visually and quantitatively. Our code is available at \url{https://amrtsg.github.io/DepGAN/}.

\end{abstract}

\input{intro}
\input{related}
\input{method}
\input{eval}
\input{conclusion}

{\small
\bibliographystyle{ieee_fullname}
\bibliography{paper}
}
\input{appendix}

\end{document}

%% file: intro.tex
\section{Introduction}
\label{sec:intro}

Image composition, which involves seamlessly integrating a foreground image with a background image to produce a unified and realistic image, has experienced significant breakthroughs with the emergence of diffusion \cite{imprint, controlcom} and GAN \cite{compgan, stgan, mtgan, patchgan} models. These models have spurred the development of generative image compositing for producing visually pleasing and realistic composites. The objective is to ensure that the foreground object in the composite image is accurately represented while blending its color and geometry seamlessly with the background. Existing methods \cite{compgan, mtgan, stgan} showcase remarkable capabilities in generative compositing; however, they frequently fall short in managing occlusion or maintaining context consistency.  

\begin{figure}[h]
  \includegraphics[width=\linewidth]{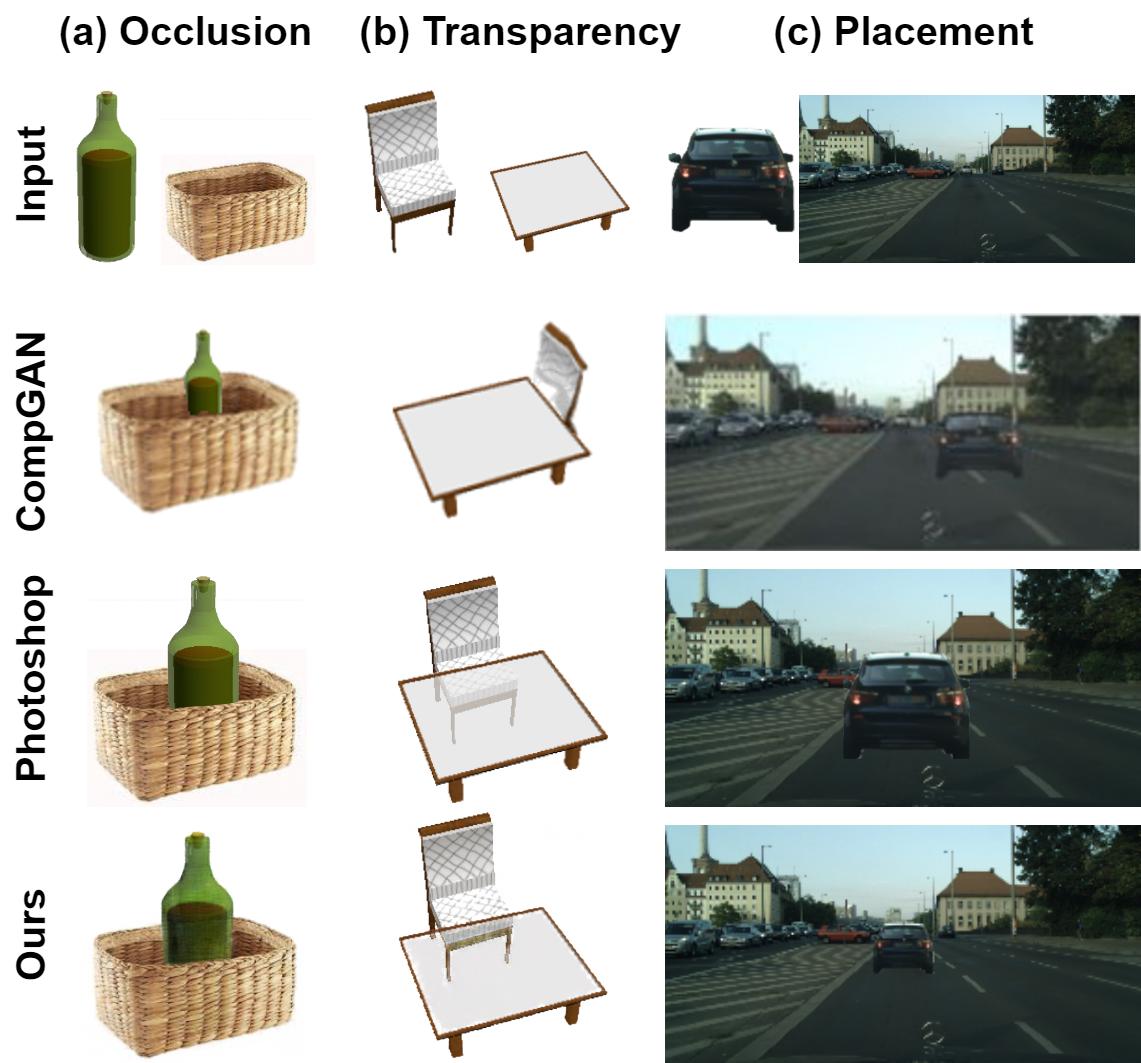}
  \caption{Comparison between CompositionalGAN \cite{compgan}, Photoshop \cite{photoshop}, and DepGAN (our work) on (a) handling occlusion, (b) handling transparency, and (c) contextual placement semantic when compositing a foreground with a background.} 
  \label{fig:composition}
\end{figure}

Occlusions occur when objects in composite image partially or fully obscure other objects, while transparency involves the blending of semi-transparent elements. For instance, Fig. \ref{fig:composition} (a) shows an example of a composition that requires occlusion handling, i.e., the bottle is partially occluded by the basket in the composite and  Fig. \ref{fig:composition} (b) shows the composition of a chair with semi-transparent table. Existing solutions for handling occlusions either need some manual intervention~\cite{partiallyoccluded} or fall short in terms of accuracy and the clear delineation of occlusion boundaries~\cite{compgan} as shown Fig. \ref{fig:compdesk}. Traditional image editing software, such as Photoshop \cite{photoshop}, often requires users to separate image contents into distinct layers and compose the image progressively from foreground to background. However, identifying and creating these layers is a complex task that demands significant manual effort and expertise. Consequently, seamless integration of a foreground into the background image, while accurately representing occlusions, remains a significant challenge in the field of image compositing.

To address this challenge, we develop an end-to-end conditional GAN (cGAN) architecture that leverages three-dimensional (3D) scene information and the alpha channel to enhance the image composites. In general, cGANs \cite{cgan} take auxiliary information such as class labels or other attributes alongside the random noise vector to generate an output image. DepGAN takes depth maps of the background images as the auxiliary information. The generator component of DepGAN is designed to create synthetic images that are virtually identical to ground truths, while simultaneously ensuring they align with their associated depth maps. The model's ability to connect depth information with characteristics of background objects leads to significant enhancements in how foreground objects are positioned within the scene and how occlusion boundaries are defined. DepGAN utilizes a standard spatial transformer network~\cite{stn} to effectively adjust the scale and position of foreground elements and utilizes alpha channel of input images to accurately render transparent/semi-transparent elements. The key contributions of this work are as follows:
\begin{itemize}[noitemsep]
    \item A versatile conditional GAN network capable of handling occlusion and transparency in image composition.
    \item A custom loss function that utilizes the information in depth maps to aid the occlusion boundary handling.
    \item A new aerial dataset of 4600 aerial images with the sole focus on contextually consistent placement of foreground objects in the scene.
\end{itemize} 

%% file: related.tex
\section{Related Work}
\label{sec:related}
Traditional image blending techniques, including alpha blending~\cite{alpha} and Poisson blending~\cite{poisson}, are essential in composite image synthesis. Alpha blending \cite{alpha} integrates images by adjusting pixel transparency based on alpha values, allowing for smooth transitions between foreground and background elements. Poisson blending, introduced by Pérez et al. \cite{poisson}, aligns the gradients of source and target images, solving a Poisson equation to achieve seamless integration. Recent advancements in image blending leverage deep learning, Neural Radiance Fields, and text-guided inpainting diffusion models to improve composite image quality and address traditional limitations \cite{deepimgblend, 3dawareblend, dreamcom}.

\paragraph{Image Harmonization}
Recent advancements in image harmonization enhance composite image realism by addressing lighting, color, and visual balance. Techniques include using lighting cues \cite{aux}, visual balance features \cite{compappeal}, fusion networks \cite{deepimagecompositing}, and color information from references \cite{cutandpaste, domainver, intrinsicimage, cutandpasteseg}. Innovative approaches involve operator masks \cite{semanticguided}, spatial attention modules \cite{spatialattention, attentionbased}, and leveraging background or foreground information \cite{backgroundguided, sidnet, foregroundaware, lemart, spatialseparated}. Advanced methods like SCS-Co address distortion by generating negative samples \cite{scsco}, while CDTNet combines color mapping with pixel transformations for high-resolution harmonization \cite{highres}, and semi-supervised strategies improve generalization from real-world images \cite{semisupervised}.

\paragraph{Image Composition}
Several surveys explore methodologies and challenges in image composition, covering object placement, semantic-aware composition, and dataset requirements \cite{compsurvey, objdata, fastobj}. Recent methods enhance image composition by decomposing processes for precise manipulation \cite{compgan}, incorporating spatial transformers for alignment \cite{stgan}, and addressing spatial placement with geometric inference \cite{singleshot}. Techniques also integrate geometric consistency \cite{geoconsistent, advertise}, spatial information \cite{sfgan}, and efficient object placement \cite{objstitch, ftopnet, mtgan, topnet, sapc}. Advanced approaches include diffusion-based generators \cite{anydoor}, exemplar-guided editing \cite{paintbyexample}, and frameworks combining blending, harmonization, and generative composition \cite{pasteinpaint, controlcom, partiallyoccluded, cutandpaste}. Recent advancements in object placement leverage context for predicting locations, with PlaceNET using inpainting \cite{placenet} and enhancing scene integration \cite{contextaware}. Techniques like masked convolutions \cite{objplacemask}, MISC framework \cite{misc}, context models \cite{importance}, and reinforcement learning \cite{interactiveobjectplacement} further improve accuracy and interactivity. In 3D information learning, Tulsiani et al. use volumetric primitives for shape abstraction \cite{lsa}, and Yuan et al. introduce CustomNet for 3D view synthesis \cite{customnet}. For object detection/segmentation, TERSE generates realistic training samples \cite{objsynth}, and Monnier et al. decompose images into object layers \cite{imgdecomp}. We refer the readers to a recent and comprehensive survey \cite{compsurvey} for more detailed insights into this domain.

%% file: method.tex
\section{Method}
In this section, we give an overview of the proposed DepGAN, and describe the network components and the loss function of DepGAN in detail.

\subsection{Objective Formulation}
We consider image datasets, $F = \{f_1, ..., f_n\}$ and $B = \{b_1, ..., b_n\}$, which contain images of foregrounds and backgrounds from their respective marginal distributions $p_F$ and $p_B$, and $C = \{c_1, ..., c_n\}$, which includes ground truth composite images containing foregrounds and backgrounds from their joint distribution $p_{data}(c)$. Additionally, we consider a set $M = \{m_1, ..., m_n\}$ of depth masks for all images in $B$ and hence follows the distribution $p_B$. For brevity, we ignore the subscripts in the following discussion. The model takes two input images $(f, b)$ and depth mask $ m $ of $b$ and generates an output image $\hat{c}$ that aims to match the target distribution $p_{data}(c)$. The model adopts cGAN~\cite{cgan} strategy, i.e., the generator ($G$) strives to create a distribution $p_g$ by utilizing $f, b,$ and $ m$ that mimics the real data distribution $p_{data}(c)$. The generator achieves this objective by transforming inputs into synthetic composites $\hat{c}$. Meanwhile, the discriminator $D$ evaluates the synthetic composites $\hat{c}$ and estimates the likelihood that $\hat{c}$ originated from the actual training data instead of the generator's distribution $p_{g}$. During the training process, the generator and discriminator are trained simultaneously using an adversarial loss function formulated as follows:

\begin{equation}
\begin{aligned}
 \min_{G} \max_{D} \mathcal{L}_{adv}(G, D) =  \mathbb{E}_{c \sim (p_{data}(c))} [\log D(c)]\\ + \mathbb{E}_{(f \sim p_{F}(f), b,m \sim p_{B}(b))} [\log (1-D(G(f, b, m)))]
\end{aligned}
\label{eq:aloss}
\end{equation}

Note that the model does not use a noise vector $z$ and as a result aims to learn a limited subset of $p_{data}(c)$ often converging to one or few modes based on $f$ and $b$. To enhance the quality of the generated images, we incorporate an additional loss term based on the $L_1$ norm which penalizes the deviation between the generated images and their corresponding ground-truth counterparts. To further enhance the realism of our composites, we incorporate a new depth-aware loss function. This additional loss term serves two key purposes: it improves the occlusion interactions between objects in the scene and enhances the rendering of transparency effects. The full model with all the loss functions can be formulated as follows:

\begin{equation}
\arg \min_{G} \max_{D} \mathcal{L}_{adv}(G, D) + \lambda_1 \mathcal{L}_{L_1}(G) + \lambda_2 \mathcal{L}_{depth}(G)
\label{eq:total_loss}
\end{equation} where \( \lambda_1 \) and \( \lambda_2 \) control the relative importance of $L_1$ and depth-aware loss terms, respectively.

\begin{figure}[h]
  \includegraphics[width=\linewidth]{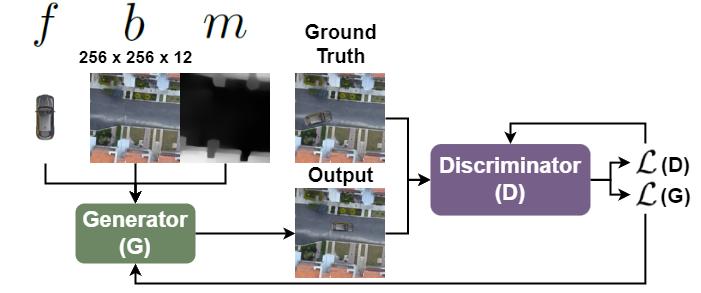}
  \caption{Overall architecture of DepGAN.}
  \label{fig:depgan}
\end{figure}

\begin{figure*}[ht]
  \includegraphics[width=\linewidth]{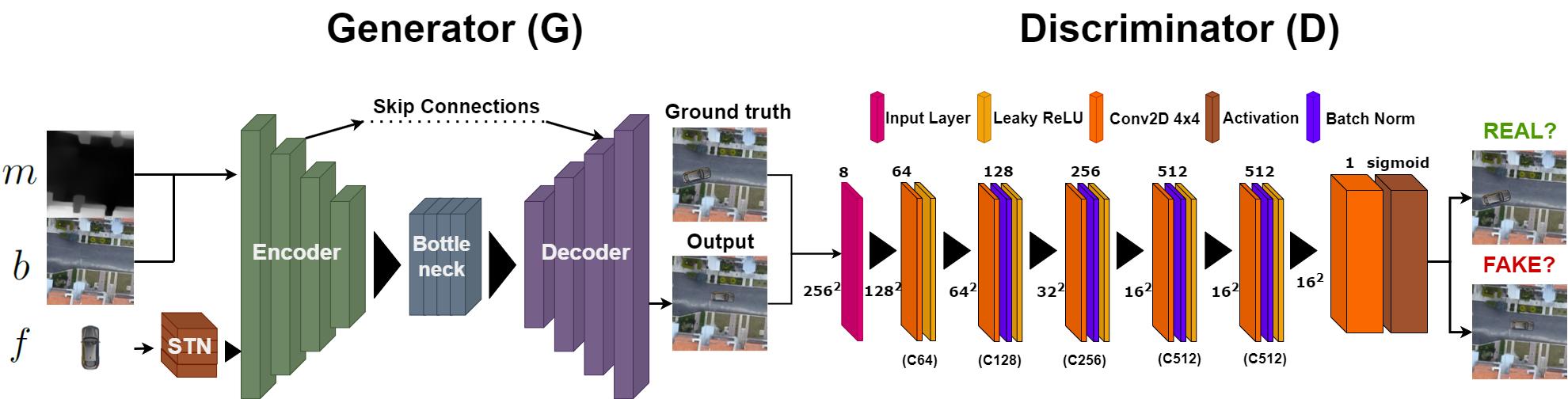}
  \caption{\textbf{Left (G):} The generator part of DepGAN. The Depth Aware Loss which penalizes the generator when generating foreground regions in composite image where the varying depth values are lighter is our addition to the network. The detailed architecture of our generator is provided in the supplementary materials. \textbf{Right (D):} A diagram of our discriminator, based off the PatchGAN discriminator \cite{patchgan}, which assesses whether each N × N patch within an image is genuine or generated. This approach allows for a finer-grained analysis of image structure.}
  \label{fig:disc}
\end{figure*}

\subsection{The Network Components}

\paragraph{Outline} Fig. \ref{fig:depgan} shows the overall architecture of DepGAN. DepGAN utilizes a generator architecture $G$ inspired by the U-Net model \cite{unet}, which is recognized for its effectiveness in image segmentation tasks. However, in the context of DepGAN, $G$ serves a different purpose; combining foreground images with background images to create composite images. DepGAN's discriminator $D$ adopts the PatchGAN \cite{patchgan} architecture, commonly used in image-to-image translation tasks. The input to the generator consists of foreground $f$, background $b$ and background depth mask $ m $ images. Foreground image undergoes a series of affine transformations and gets aligned well with the background \( b \). The transformed foreground \( f^{'}  \) is then concatenated with \( b \) and \( d \). The new input goes through the generator layers to generate the predicted image $\hat{c}$. We then concatenate the output $\hat{c}$ with the ground truth $c$, and feed it to $D$ which tries to distinguish between both the images, and attempts to determine whether the image generated by $G$ $\hat{c}$  is fake or not. Note that each input, output and ground truth images have a spatial dimension of $256 \times 256$ with four channels including color (R,G,B) and alpha channels, i.e., $I \in \mathbb{R}^{256 \times 256 \times 4}$. 
\paragraph{Generator} The generator architecture is tailored specifically for image composition tasks with additional conditional input, i.e., depth maps. Conditional GANs have limitations when it comes to spatial object transformations. To address this, we explicitly incorporate scale and shift transformations by utilizing a Spatial Transformer Network (STN)\cite{stn} to transform the foreground image. This approach allows for more precise control over the spatial aspects of object manipulation. The U-Net part of the generator consists of an encoder, a bottleneck, and a decoder, with skip connections connecting each encoder block with its corresponding decoder block, refer to Fig. \ref{fig:disc}. Skip connections facilitate the direct flow of information from the encoder layers (low-level features) to the decoder layers (high-level features). This helps in preserving pixel-wise spatial information that might get lost in deeper layers of the network. As a result, the network can better localize and refine details in the reconstructed image. Specific details of the architectures are provided in the supplementary materials. We use $L_1$ loss, for image reconstruction in the generator which acts as a corrective measure, pushing the model to generate images that faithfully represent their ground-truth originals. The loss can be formulated as follows:
\begin{equation}
\label{eq:mae_loss}
\mathcal{L}_{L{_1}} = \frac{1}{N} \sum_{i=1}^{N}  |(c_i - \hat{c_i})| 
\end{equation}
Here, \(N\) is the total number of pixels.

\paragraph{Discriminator}
We use the PatchGAN discriminator framework \cite{patchgan} (Fig. \ref{fig:disc}) in DepGAN, due to its impressive pixel-wise accuracy in distinguishing between generated and ground truth images. The discriminator operates by discerning between images at a local level, evaluating the difference between individual patches rather than assessing images as a whole. By evaluating image patches instead of entire images, the PatchGAN discriminator provides detailed feedback during adversarial training, enabling more precise discrimination between real and generated composite images. The architecture is made up of multiple convolutional layers, each progressively extracting hierarchical features from the concatenated input images. The convolutional layers, labeled C64, C128, C256, C512, and optionally, another C512 increase in complexity and filter count, capturing increasingly detailed aspects of the input images. 

\subsection{Depth Aware Loss (DAL)}
\label{sec:dalloss}

\begin{figure}[h]
\centering
  \includegraphics[width=\linewidth]{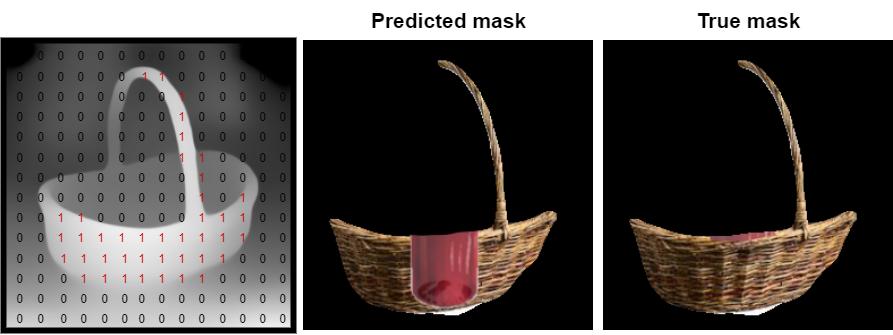}
  \caption{\textbf{Left.} An image of the binary mask where each foreground pixel imposes a penalty or reward on the generator. Lighter pixels indicate higher penalties. The depth mask applied to the predicted image and ground truth highlights only important areas, ensuring the foreground does not overlap the background.}
  \label{fig:dal}
\end{figure}

Traditional approaches to evaluating the quality of generated composite images often rely solely on pixel-wise differences between the predicted and ground truth images. However, these methods quite often overlook critical factors such as depth information, which plays a pivotal role in determining the spatial relationships and visual coherence of composite scenes. To address this issue, we utilize depth aware loss which considers depth information alongside the input images, enabling the generator to produce outputs that accurately reflect the structural characteristics of the scene. By penalizing inconsistencies in regions with varying depths, the depth aware loss function ensures that the generated images maintain spatial coherence and depth consistency, particularly in areas where foreground objects interact with the background environment. Essentially, we penalize the model for rendering the foreground in lighter depth regions using a depth penalty, i.e., the absolute difference between the masked predicted $c$ and masked ground truth $\hat{c}$ images, multiplied by the depth mask as shown below:

\begin{equation} \label{eq:depth_loss}
\mathcal{L_\text{depth}} = \frac{1}{N} \sum_{i=1}^{N} |(c_i - \hat{c_i})| \cdot \chi(m_i)\end{equation}Where $\mathcal{L_{\text{depth}}}$ is the depth-aware loss, \(N\) is the total number of pixels, $c$ is the predicted image, and $\hat{c}$ is the ground truth image. We use an indicator function $\chi(m_i)$ to distinguish foreground and background regions. Firstly, we extract the depth map from the ground truth image, and create binary mask based on a threshold value. Depth pixels with values above the threshold are set to 1, and values below are set to 0 as shown in Fig. \ref{fig:dal} and the equation below:
\begin{equation} \label{eq:chi}
  \chi(m_i) =
    \begin{cases}
      1 & \text{if} \ m_i \geq threshold \\
      0 & \text{otherwise}
    \end{cases}       
\end{equation}

The threshold value varies from dataset to dataset, we set this value to an optimal line between the foreground and background manually depending on the dataset used.  This mask is then applied to the predicted and ground truth images, effectively zeroing out the regions where the depth is below the threshold. Fig. \ref{fig:dal} shows the result of this mask application. DAL selectively emphasizes errors in foreground regions (where depth values are higher), while mitigating the impact of discrepancies in background regions (where depth values are lower). This ensures that the model prioritizes accuracy in regions with significant depth variation, resulting in composite images that maintain depth consistency and visual coherence. 

%% file: eval.tex
\section{Experiments}
\label{sec:eval}
We evaluate DepGAN through quantitative and qualitative evaluations on synthetic and real datasets. In the evaluations, we focused on analyzing the effectiveness of DepGAN to the inherent challenges posed by occlusion and transparency in image composition tasks. We compared our results with state-of-the-art methods such as Compositional GAN and ST-GAN \cite{compgan, stgan} due to their established effectiveness in handling similar challenges. For qualitative analysis, we included comparisons with Photoshop \cite{photoshop}, as it is considered the gold standard for manual image composition. However, we omitted Photoshop from our quantitative analysis due to the manual interventions required. We used the \textit{Structural Similarity Index (SSIM)} that measures similarity by comparing structural information in images with higher scores indicating better similarity), and the \textit{Peak Signal-to-Noise Ratio (PSNR)}, which assesses image quality by comparing the peak signal power and noise level with higher values indicating lower distortion. Additionally, we utilize \textit{Mean Absolute Error (MAE)} that measures the average absolute difference between the generated and ground truth images, with lower values indicating better accuracy and the \textit{Perceptual Similarity Metric (LPIPS)} \cite{lpips} uses features from deep neural networks like AlexNet or VGG (trained on ImageNet dataset \cite{imagenet}) to capture human perception of image similarity, with smaller distances indicating higher perceptual similarity.
\subsection{Evaluation on Real-World Datasets}
\begin{figure}[h]
\centering
\includegraphics[width=\linewidth]{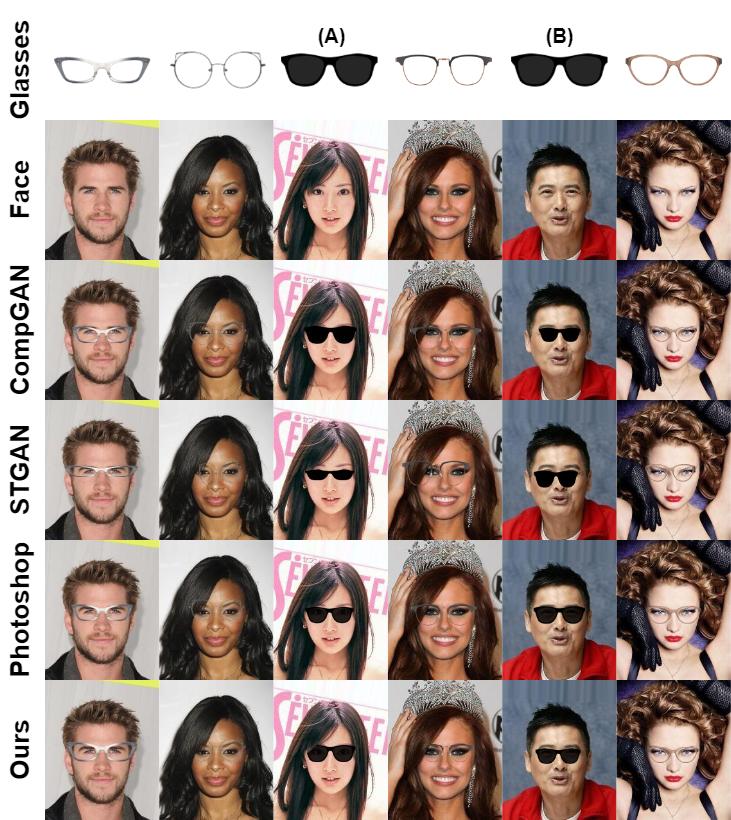}
\caption{Evaluation on STRAT’s dataset \cite{STRAT}: Both DepGAN and Compositional GAN show impressive performance, but in cases marked (A) and (B), only DepGAN is able to handle the transparency of the glasses.}
\label{fig:glasscomp}\end{figure}

\paragraph{STRAT's Dataset~\cite{STRAT}.} We tested DepGAN on real-world datasets comprising authentic scenes and objects. The real-world datasets include STRAT's \cite{STRAT} dataset (face-glasses), and the car-street dataset used by Compositional GAN \cite{compgan}. In our evaluation on STRAT's dataset, DepGAN was the only model which successfully handled occlusion and transparency, hiding the glasses behind the hair where necessary, as shown in Fig. \ref{fig:glasscomp}. This is  an advantage of using depth information in our model. Transparent cases can be seen in rows marked (A) and (B) in Fig. \ref{fig:glasscomp}, resulting from the inclusion of the alpha channel during model training. Table \ref{tab:glassesacc} indicates that DepGAN shows the highest accuracy between the three models, one reason for the increased accuracy is due to the inclusion of transparency in the glasses within the ground truth images. Since manual editing is involved in Photoshop results, we exclude it from the quantitative comparison.

\begin{table}[h]
\centering
\caption{A quantitative comparison on STRAT's face-glasses dataset \cite{STRAT}. DepGAN outperformed ST-GAN and Compositional GAN in both quantitative and qualitative assessments, being the only model to handle occlusion and transparency.}
\small
\begin{tabular}{lllll}
\hline
\textbf{Model} & \textbf{MAE}$\downarrow$ & \textbf{PSNR}$\uparrow$ & \textbf{SSIM}$\uparrow$ & \textbf{LPIPS}$\downarrow$ \\ \hline
CompGAN\cite{compgan} & 1.54  & 29.26   & 0.96    & 0.02  \\ 
STGAN\cite{stgan}            & 2.09  & 28.31   & 0.95    & 0.04  \\ 
DepGAN            & \textbf{1.17}  & \textbf{31.25}   & \textbf{0.97}    & \textbf{0.01}  \\ \hline
\end{tabular}
\label{tab:glassesacc}
\end{table}

\begin{figure}[h]
\includegraphics[width=\linewidth]{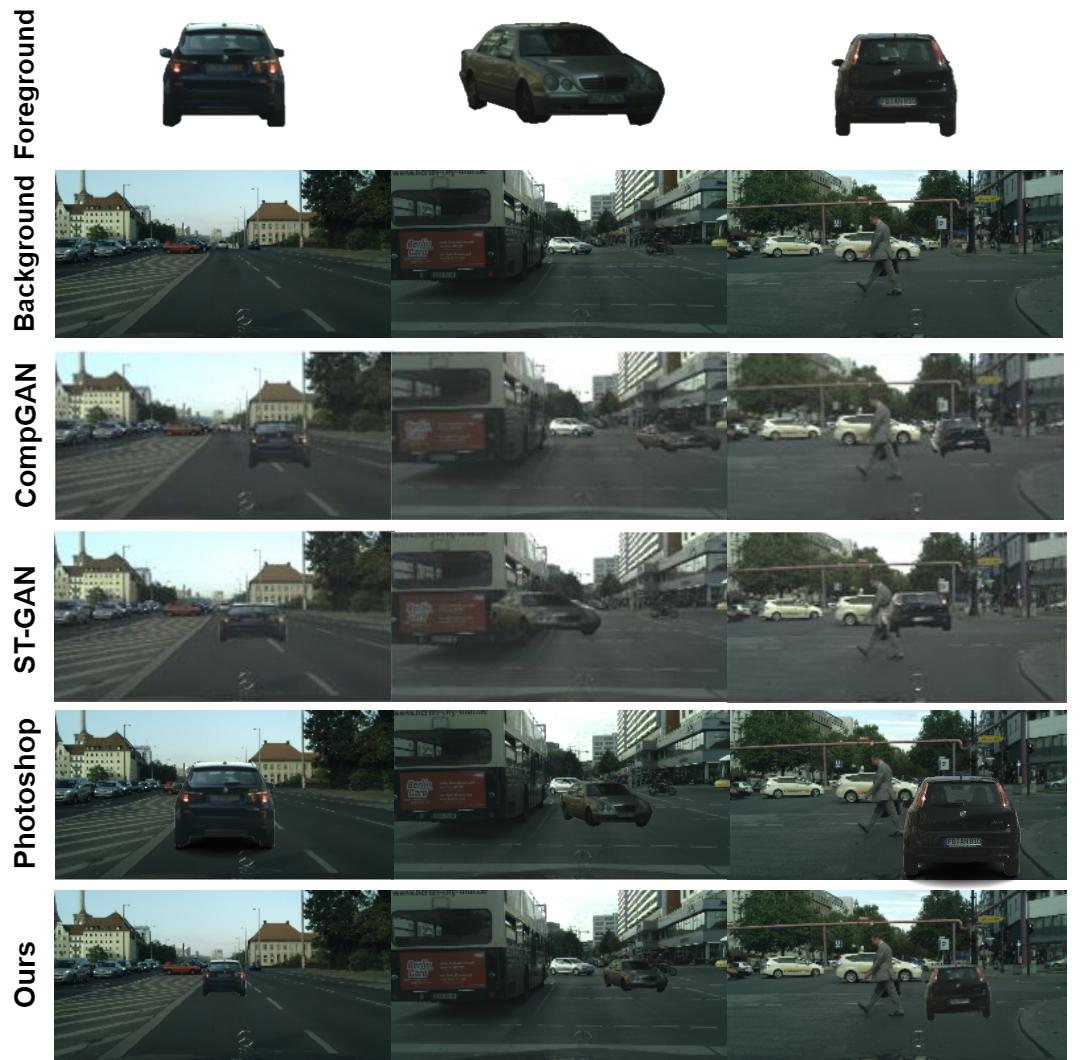}
\caption{In our comparison of results obtained from the car-street dataset~\cite{compgan}, DepGAN shows a superior scaling and placement of the cars in relation to their surroundings within the street scene compared to the outputs produced by the other models.}
\label{fig:realcarcomp}\end{figure}
\paragraph{Car-Street Dataset~\cite{compgan}.}The second experiment was done on the car-street dataset. In this test, we remove the foreground image (car) from one image and composite it with a different background image. The primary goal of this experiment is to assess the model's ability to align the perspective of the foreground with the background and accurately place it in the correct area, the results can be seen in Fig. \ref{fig:realcarcomp}. Qualitatively, DepGAN's and Compositional GAN's results are more accurate in terms of placement and perspective.

\subsection{Evaluation on Synthetic Datasets}
\begin{figure}[h]
\includegraphics[width=\linewidth]{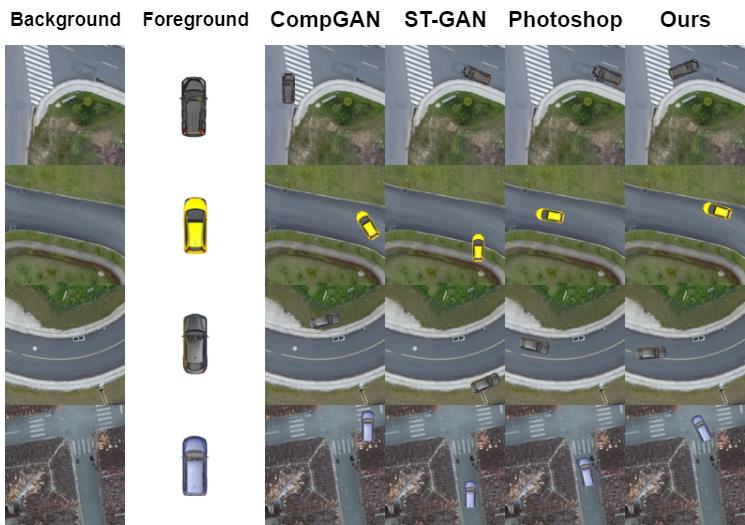}
\caption{A comparison of results on our aerial dataset between Compositional GAN, ST-GAN and DepGAN (ours). Out of the test set, DepGAN performed the best when placing the car on the road.}
\label{fig:aerialcomp}\end{figure}

\paragraph{Aerial Dataset.} We synthesized an aerial image dataset comprising 4600 images. The dataset consists of aerial views of cars as foregrounds and aerial background images captured from drones traversing varied landscapes (More details in the supplementary materials). We test a model's semantic placement performance by quantifying how accurately it can place the foreground (car) in the correct location in the background scene. In our context, i.e., cars and road scenes, as long as the whole car is in the street, we count it as a correct placement. In our evaluation, we provided each model with the same test dataset consisting of 50 image pairs. Example results can be seen in Fig. \ref{fig:aerialcomp}. Out of 50 images, ST-GAN was able to correctly place the car in 36 composites, Compositional GAN in 24, and DepGAN in 41, as shown in Table \ref{tab:aerialcomp}. We gathered the average MAE, PSNR and SSIM scores of all 50 images for each model. ST-GAN and DepGAN outperformed Compositional GAN with similar scores (refer to Table \ref{tab:aerialacc}) potentially due to the accurate placement of the foregrounds much similar to the respective ground truths.

\begin{table}[h]
\centering
\caption{A comparison between each model on our aerial dataset, a correct placement is counted if the whole foreground (car) is inside the bounds of the street. DepGAN performed the best with 41 correct placements out of 50.}
\small
\begin{tabular}{lll}
\hline
\textbf{Model}             & \textbf{Correct} & \textbf{Incorrect} \\ \hline
CompGAN \cite{compgan} & 24  & 26  \\ 
STGAN \cite{stgan}  & 36  & 14\\ 
DepGAN  & \textbf{41}   & \textbf{9}  \\ \hline
\end{tabular}
\label{tab:aerialcomp}
\end{table}
\begin{table}[h]
\centering
\caption{A quantitative comparison between DepGAN, ST-GAN, and Compositional GAN on our aerial dataset.}
\small
\begin{tabular}{lllll}
\hline
\textbf{Model}    & \textbf{MAE}$\downarrow$ & \textbf{PSNR}$\uparrow$ & \textbf{SSIM}$\uparrow$ & \textbf{LPIPS}$\downarrow$\\ \hline
CompGAN \cite{compgan} & 2.75  & 25.42   & 0.93  & 0.09    \\ 
STGAN \cite{stgan} & \textbf{2.31}  & 26.07   & \textbf{0.95}  & 0.07    \\ 
DepGAN   & \textbf{2.31}  & \textbf{26.73}   & 0.94  & \textbf{0.06}    \\ \hline
\end{tabular}
\label{tab:aerialacc}
\end{table}
\begin{figure}[h]
\includegraphics[width=\linewidth]{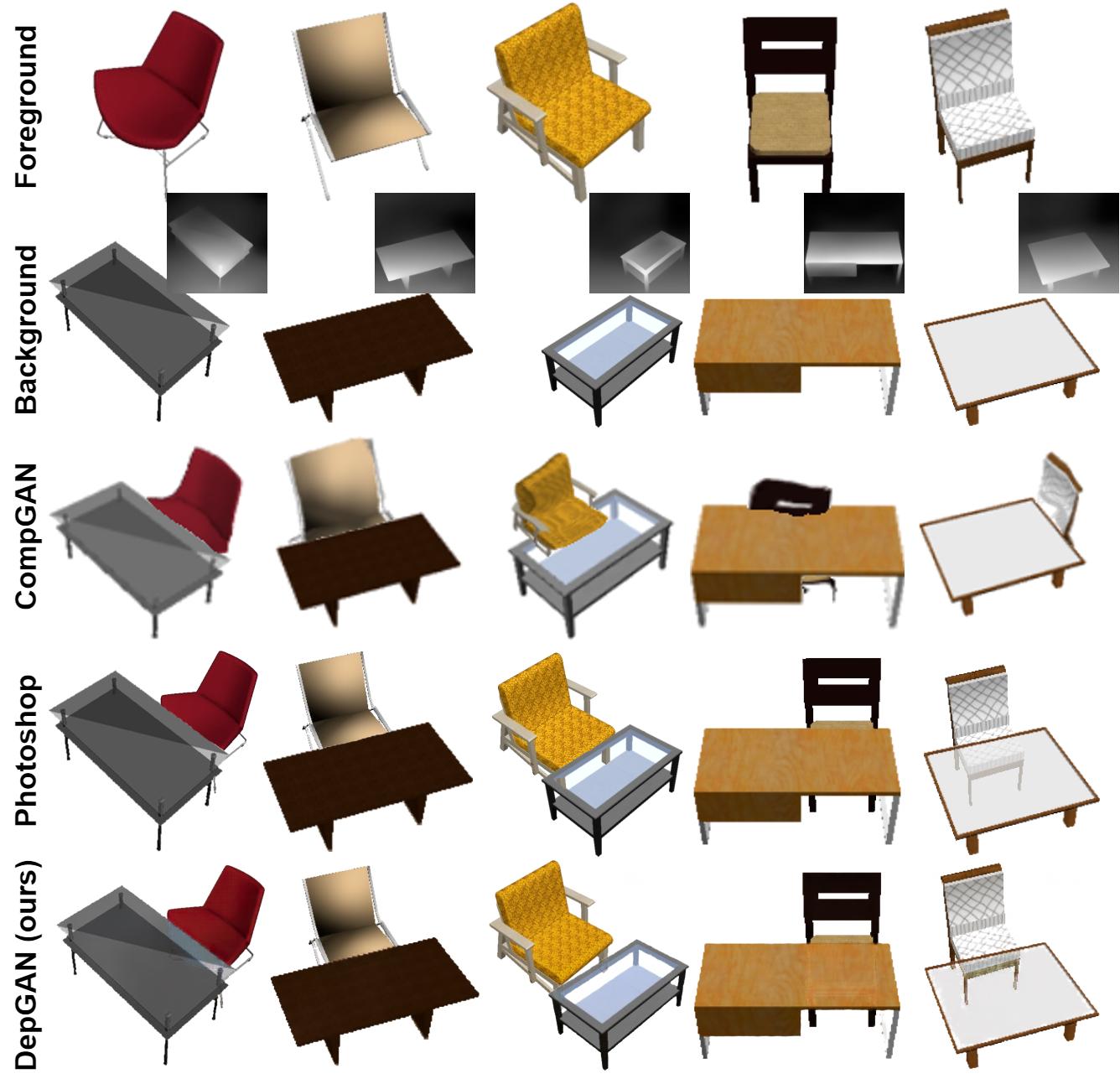}
\caption{Qualitative results on the chair-desk dataset. DepGAN's results demonstrate clear and precise occlusion of the chair from the desk. In contrast, Compositional GAN struggles to accurately delineate the occlusion boundary, resulting in less defined outlines.}
\label{fig:compdesk}\end{figure}

\begin{figure}[h]
\includegraphics[width=\linewidth]{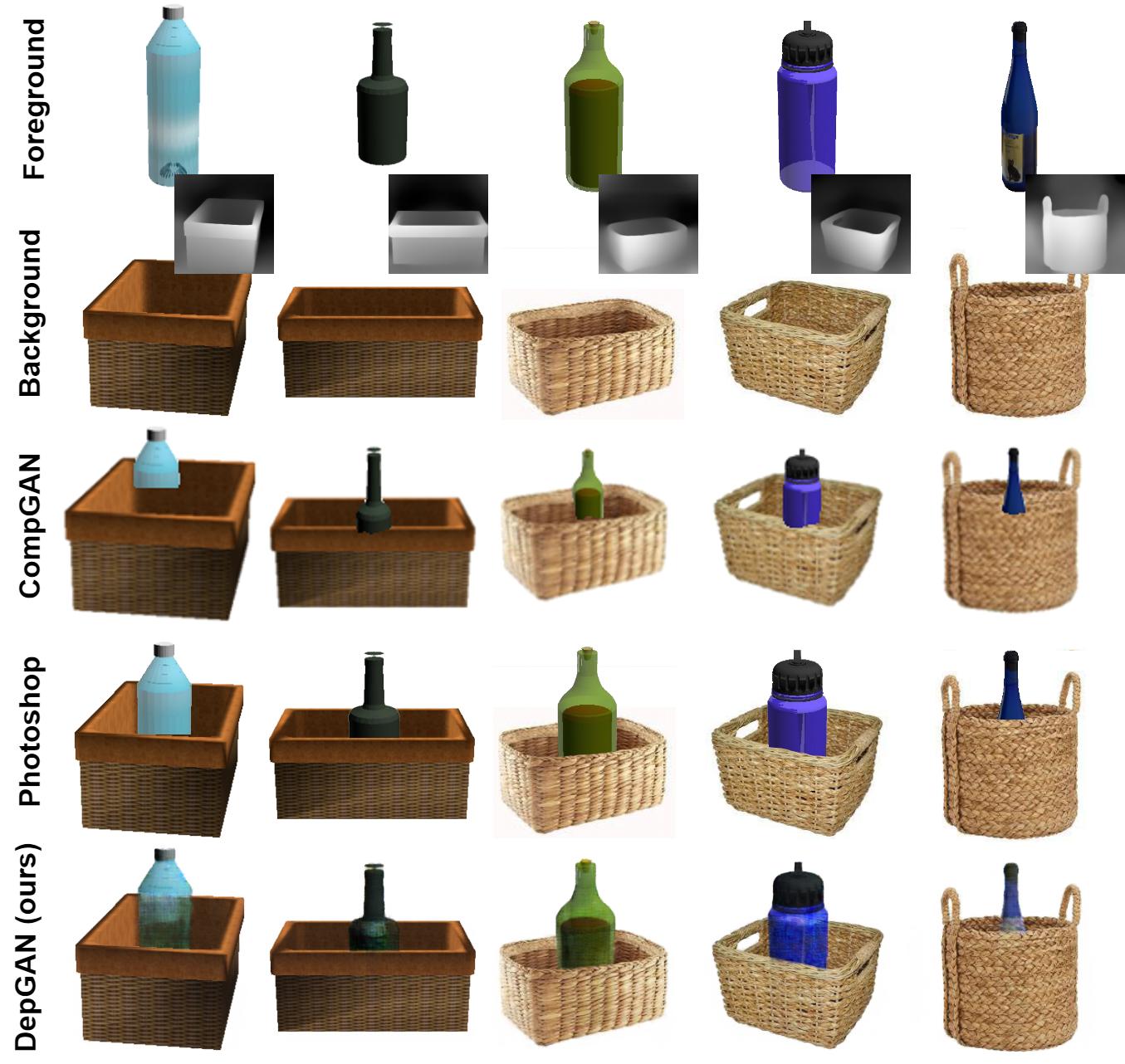}
\caption{Comparison of results on the bottle-basket dataset between different methods. In DepGAN's results, the delineation of the occlusion boundary for the bottle is sharp and clear, while Compositional GAN fails to properly identify the boundary of the basket for occlusion.}
\label{fig:compbottle}\end{figure}

\paragraph{Shapenet Datatset\cite{shapenet}.}The second synthetic dataset we tested DepGAN on is Shapenet's dataset\cite{shapenet}, which consisted of 2 categories each providing different examples of occlusion. The first dataset included images of chairs and desks/tables (Fig. \ref{fig:compdesk}). The second dataset included foreground images of bottles and background images of baskets (Fig. \ref{fig:compbottle}). The sole focus of this experiment was to test the model's performance on occlusion handling capability. Ideally, the bottle should be placed inside the basket and the chair should be placed behind the desk/table. Quantitatively, DepGAN yields more accurate scores, refer to Table \ref{tab:bottle-scores}. DepGAN's accurate metric scores validates its ability to correctly delineate the occlusion boundaries, keeping the background area as close to the ground truth as possible. Table \ref{tab:bottle-scores} indicates that Compositional GAN struggles to position the chairs correctly which directly contributed to its lower scores, while in some cases, failing to handle the transparency found in the ground truth images. Qualitatively, DepGAN outperforms Compositional GAN when it comes to the occlusion delineation. DepGAN's results show a more accurate delineation of the occlusion boundaries with the line between the basket and bottle being sharp and precise as shown in Fig. \ref{fig:compbottle}. When drawing the foreground itself, Compositional GAN does a better job. Although DepGAN does generate the foreground well, there are some artifacts. In the case of the chair/desk dataset, artifacts did not seem to be an issue, with DepGAN performing considerably better than Compositional GAN as shown in Fig. \ref{fig:compdesk}.

\begin{table}[h]
\centering
\caption{A quantitative comparison of different models on Shapenet\cite{shapenet} datasets.}
\scriptsize
\begin{tabular}{cccccc}
\hline
\textbf{Category} &\textbf{Network} & \textbf{MAE}$\downarrow$ & \textbf{PSNR}$\uparrow$ & \textbf{SSIM}$\uparrow$ & \textbf{LPIPS}$\downarrow$\\
\hline
\multirow{2}{*}{Chair-Table}&CompGAN & 34.62 & 11.3 & 0.65 & 0.31\\
  &DepGAN & \textbf{2.71} & \textbf{30.73} & \textbf{0.97} & \textbf{0.09}\\
\hline
\multirow{2}{*}{Bottle-Basket} & CompGAN & 24.79 & 14.51 & 0.50 & 0.19\\
&DepGAN & \textbf{3.23} & \textbf{30.63} & \textbf{0.96} & \textbf{0.07}\\
\hline
\end{tabular}
\label{tab:bottle-scores}
\end{table}

\subsection{Hyper Parameter Setting}
\begin{figure}[h]
\includegraphics[width=\linewidth]{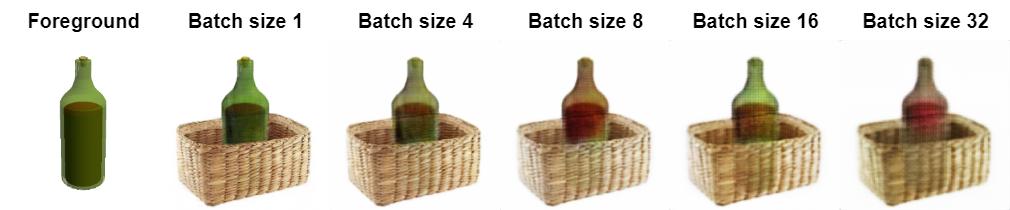}
\caption{A comparison between different batch sizes, we found that using a batch size of 1 is best for image composition tasks}
\label{fig:batch}\end{figure}

\paragraph{Effects of Batch Size.} We initially experimented with varying batch sizes while keeping other hyperparameters constant. In our experiments, a batch size of 1 significantly improved the visual quality of composite images. To analyze this, we examined accuracy and loss curves for different batch sizes, used quantitative metrics to evaluate image quality (Table \ref{tab:mae-scores}) and qualitative results as shown in Fig. \ref{fig:batch}. While this is contrary to the usual preference for larger batch sizes to speed up convergence, we observe that processing of individual images avoids the gradient averaging across multiple samples. Consequently, the composites preserve the image specific nuances such as sharper edges, and better texture (see Figure \ref{fig:batch}). This experiment reiterates that the increased noise in gradient estimates with a smaller batch size helps escape sharp local minima, whereas larger batch sizes offer smoother gradients but may lead to suboptimal solutions.

\begin{table}[h]
\centering
\caption{MAE scores for different batch sizes, results show using a batch size of 1 yields the best results.}
\small
\begin{tabular}{cccccc}
\hline
\textbf{Batch Size} & 1 & 4 & 8 & 16 & 32 \\
\hline
\textbf{MAE Score}$\downarrow$ & \textbf{2.79}& 7.79& 5.01& 6.02& 15.60 \\
\hline
\end{tabular}
\label{tab:mae-scores}
\end{table}

\begin{figure}[h]
\includegraphics[width=\linewidth]{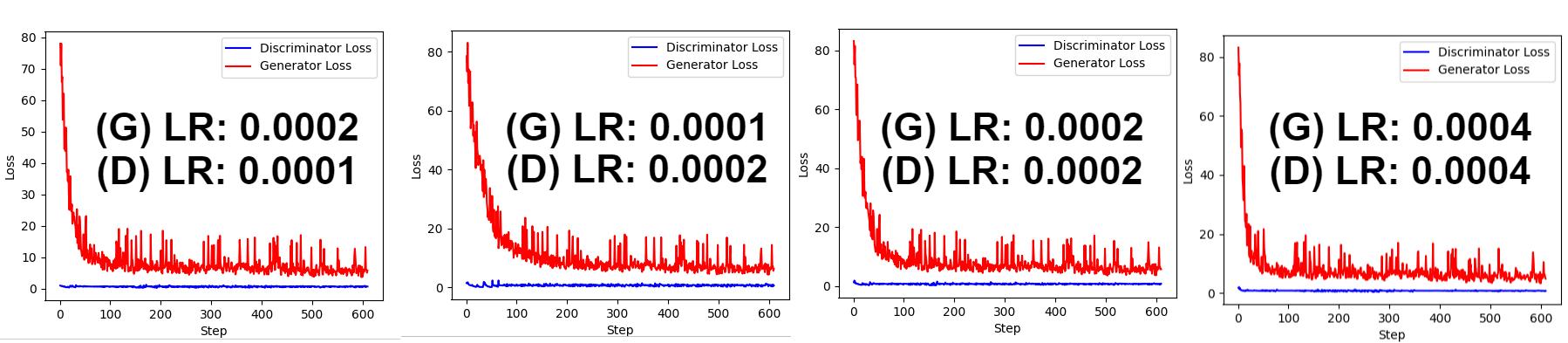}
\caption{A comparison between different learning rates.}
\label{fig:lr}
\end{figure}

\paragraph{Addressing Mode Collapse.} In our experiments, we focused on performance metrics to evaluate different learning rate configurations on Shapenet's datasets \cite{shapenet} (refer to Fig. \ref{fig:lr}), particularly mode collapse in GANs where the generator fails to capture data diversity \cite{wgan}. Mode collapse occurred when the discriminator outperformed the generator, limiting sample diversity. We identified a learning rate configuration of the generator at 0.0002 and the discriminator at 0.0001 for the Shapenet dataset. This setup reduced mode collapse, improved training stability, sample accuracy, and convergence speed, as shown in Table \ref{tab:lr-qual}. 


\begin{table}[h]
\small
\centering
\caption{Effect of learning rates on GAN training performance, in order of best quality to worst.}
\begin{tabular}{llll}
\hline
\textbf{LR} & \textbf{MAE}$\downarrow$ & \textbf{Collapse} & \textbf{Convg. (min)} \\ \hline
G 0.0002 D 0.0001 & 2.32 & no & 31 \\ 
G 0.0002 D 0.0002 & 2.98 & no & 32 \\ 
G 0.0001 D 0.0001 & 3.87 & no & 37 \\ 
G 0.0001 D 0.0002 & 4.22 & no & 48 \\ 
G 0.0003 D 0.0002 & 5.14 & no & 28 \\ 
G 0.0003 D 0.0003 & 17.63 & yes & Collapsed \\ 
G 0.0004 D 0.0004 & 28.76 & yes & Collapsed \\ \hline
\end{tabular}
\label{tab:lr-qual}
\end{table}

\subsection{Ablation on Depth-Aware Loss}
We begin by establishing a baseline model. We create a variant of the model which does not include the depth aware loss. To ensure fairness in comparison, both versions of the model are trained on the basket-bottle dataset, and is divided into training, validation, and test sets, with consistent distribution across both models.

\begin{figure}[h]
\centering
\includegraphics[width=\linewidth]{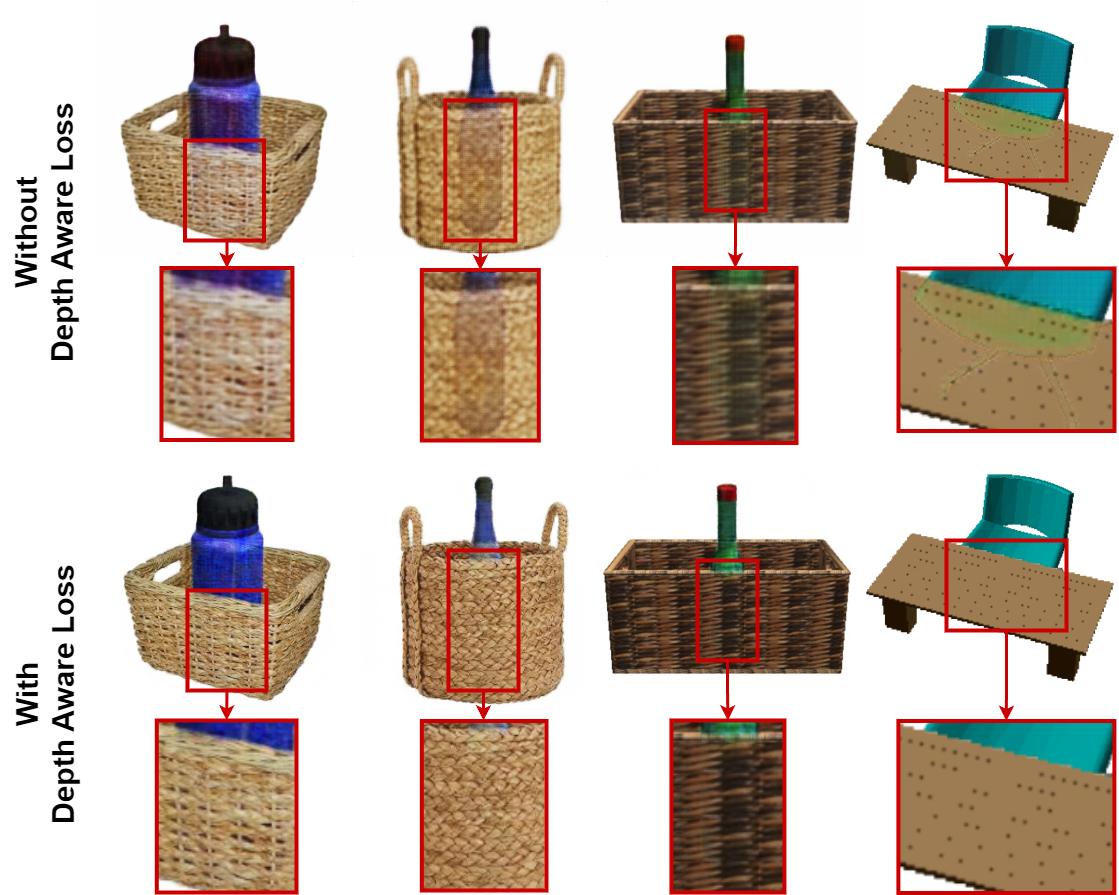}
\caption{In a qualitative experiment conducted without the Depth Aware Loss vs with Depth Aware Loss, noticeable inconsistencies arise in the rendering of the foreground, particularly in areas where it should not be present.}
\label{fig:ablation}\end{figure}

The results of the ablation studies reveal significant differences in the performance of the two models. Quantitatively, the model with the Depth Aware loss consistently outperforms the variant without it across all evaluation metrics, as shown in Table \ref{tab:ablation-scores}. Qualitatively, the inclusion of the Depth Aware Loss significantly enhances the 3D awareness of the model. This improvement enables the model to distinctly outline boundaries where occlusion of the bottle should occur as evident in Fig. \ref{fig:ablation}. Without this component, the model struggles to accurately perceive depth and outline occlusion boundaries and generates subtle outlines of the bottle where it should be occluded.

\begin{table}[ht]
\centering
\caption{\textbf{Ablation on DAL:} Comparison across metrics on DepGAN with and without Depth Aware Loss (DAL), the inclusion of DAL significantly enhanced the model's performance.}
\begin{tabular}{lllll}
\hline
\textbf{Mode} & \textbf{MAE}$\downarrow$ & \textbf{PSNR}$\uparrow$ & \textbf{SSIM}$\uparrow$ & \textbf{LPIPS}$\downarrow$\\
\hline
Without DAL & 9.93 & 21.31 & 0.73 & 0.11\\
With DAL & \textbf{3.23} & \textbf{30.63} & \textbf{0.96} & \textbf{0.07}\\
\hline
\end{tabular}
\label{tab:ablation-scores}
\end{table} 

%% file: conclusion.tex
\begin{figure}[ht]
\centering
\includegraphics[width=5.0 cm]{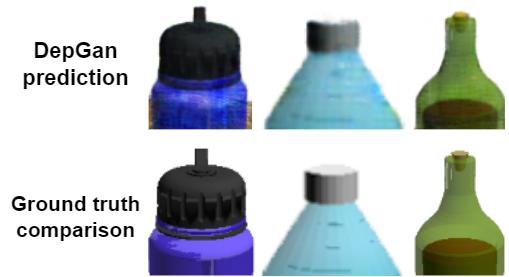}
\caption{A comparison between the rendered foregrounds generated by DepGAN and the ground truth foregrounds, highlighting artifacts present in the DepGAN output.}
\label{fig:artifacts}\end{figure}

\section{Conclusion}
We propose a GAN for image composition using depth maps. DepGAN's evaluation across synthetic and real-world datasets, alongside training and tuning processes, provides insights into its effectiveness in handling occlusion and transparency in image composition. While showing adaptability and robustness, it faces challenges such as artifacts in rendered foregrounds due to its nature as a translation model and the complexity of foreground objects as shown in Fig. \ref{fig:artifacts}. Strategies for artifact reduction include enhancing texture preservation and improving contextual understanding through scene parsing. Future research should focus on refining depth-aware loss, integrating multi-modal inputs and enhancing training strategies by considering Wasserstein loss \cite{wgan}.

%% file: appendix.tex
\clearpage
\setcounter{section}{0} 
\renewcommand*{\thesection}{\Alph{section}}
\maketitlesupplementary

\section{Aerial Image Dataset for Image Composition}
\label{sec:aerields}
\begin{figure}[h]
\centering
  \includegraphics[width=\linewidth]{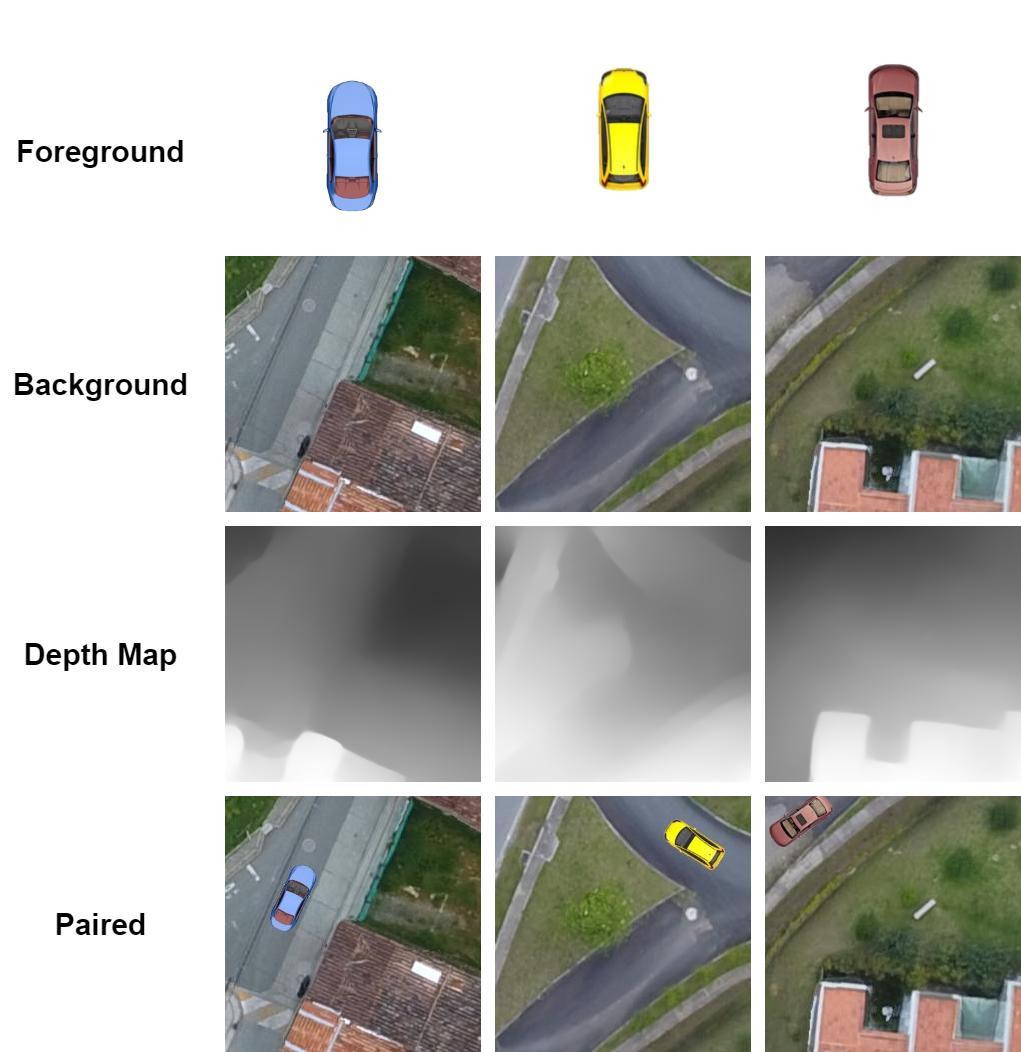}
  \caption{Three examples from our aerial car-street dataset, composed of 4600 of foreground, background, depth map, and paired images.}
  \label{fig:aerial}
\end{figure}

In our study, we found a lack of diverse and sufficient training datasets for image composition tasks. For instance, to the best of our knowledge,  no dataset exists in public domain that deals with the composition of aerial images and foreground objects. So, to test the efficacy of our model on such images, we synthesized an aerial image dataset comprising 4600 images specifically designed to train machine learning models in image composition tasks as shown in Fig. \ref{fig:aerial}.

A distinctive feature of our dataset is its deliberate focus on training machine learning models in semantic object placement within composite images. Through careful curation, the dataset aims to enhance models' abilities to intelligently place objects within scenes, emphasizing spatial relationships and semantic coherence.

Notably, our dataset includes aerial view backgrounds captured from drones traversing varied landscapes. These aerial perspectives offer unique vantage points rarely found in conventional datasets, providing valuable spatial context and depth. This characteristic enriches the dataset with diverse perspectives, challenging models to adeptly handle unique viewpoints and enhancing the realism of composite images.

The dataset was created through a systematic synthesis process to ensure high-quality training samples for machine learning models.

\begin{enumerate}
    \item \textbf{Foreground Image Acquisition:} Foreground images, depicting top views of cars, were sourced from royalty free 3D models \cite{free3d}. This ensured a diverse selection of vehicle models and configurations, enriching the dataset.
    
    \item \textbf{Background Image Selection:} Aerial background images were sourced from a curated drone road dataset \cite{roadds}, capturing diverse landscapes. These images provide a rich backdrop for composite images.
    
    \item \textbf{Depth Map Generation:} MiDaS  \cite{midas}, was used to generate depth maps for each background image. These depth maps provide spatial context for realistic integration of foreground and background elements.
    
    \item \textbf{Pairing Foreground and Background Images:} Each background image was paired with a corresponding foreground image using photoshop, ensuring accurate semantic placements, such as placing a car on the road, and not somewhere where it does not belong, i.e., on top of a roof.
\end{enumerate}

Refer to Table \ref{tab:dataset} for the distribution of images in the train-test-validation split of the dataset. We used our dataset to compare and analyze ST-GAN \cite{stgan}, Compositional GAN \cite{compgan}, Adobe Photoshop and DepGAN (the proposed) in Section 4.2 of the main paper.

\begin{table}[h]
\centering
\caption{Number of images of each sample in train-test-validation split.}
\begin{tabular}{ll}
\hline
\textbf{Split}  & \textbf{Number of Images}\\ \hline
Train      & 975 \\ \hline
Test       & 75  \\ \hline
Validation & 75  \\ \hline
\end{tabular}
\label{tab:dataset}
\end{table}

\begin{figure*}[ht]
\centering
  \includegraphics[width=15cm]{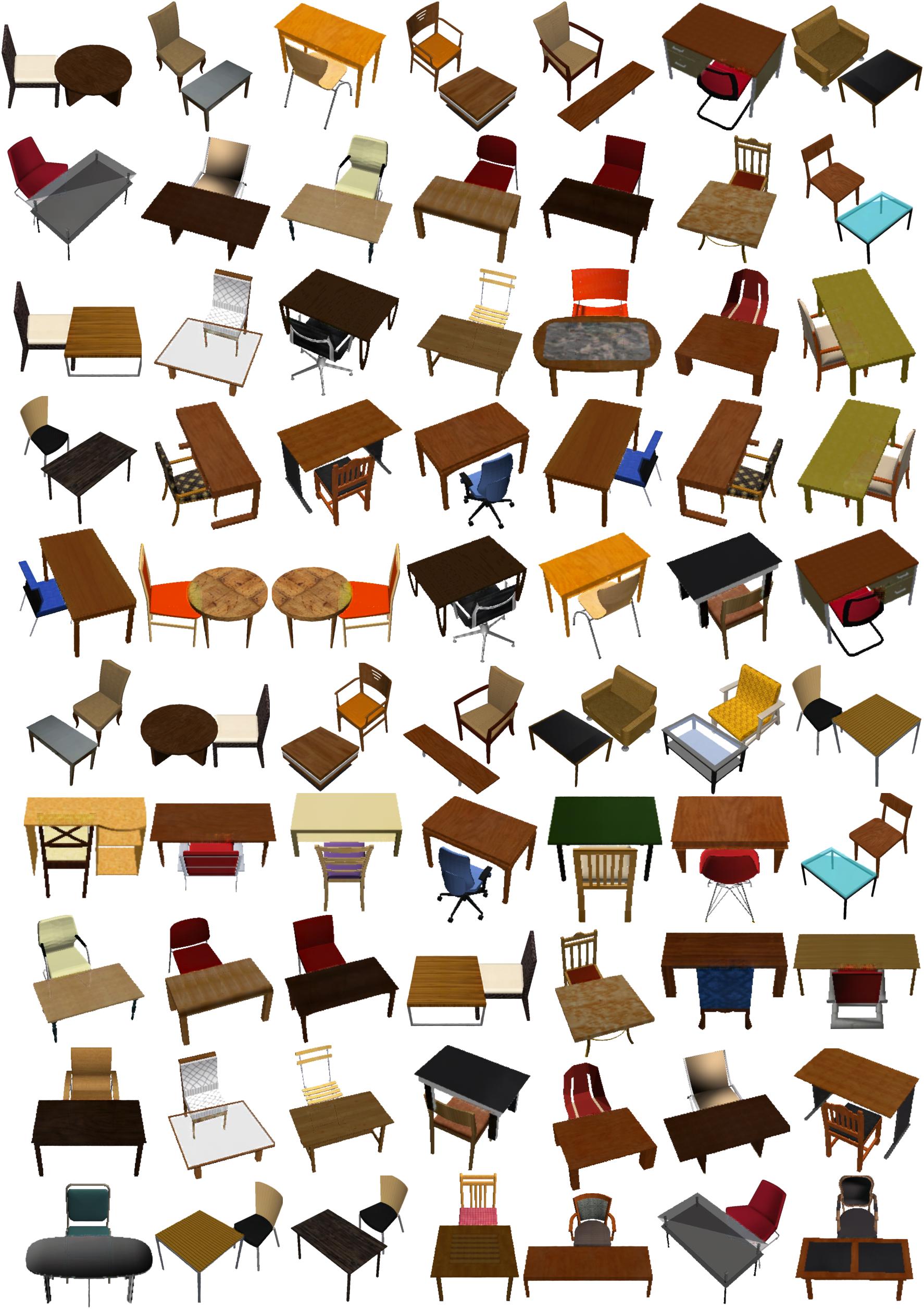}
  \caption{More results generated by the proposed network (DepGAN) on the chair/desk shapenet\cite{shapenet} dataset.}
  \label{fig:resultsfulldesk}
\end{figure*}
\begin{figure*}[ht]
\centering
  \includegraphics[width=10cm]{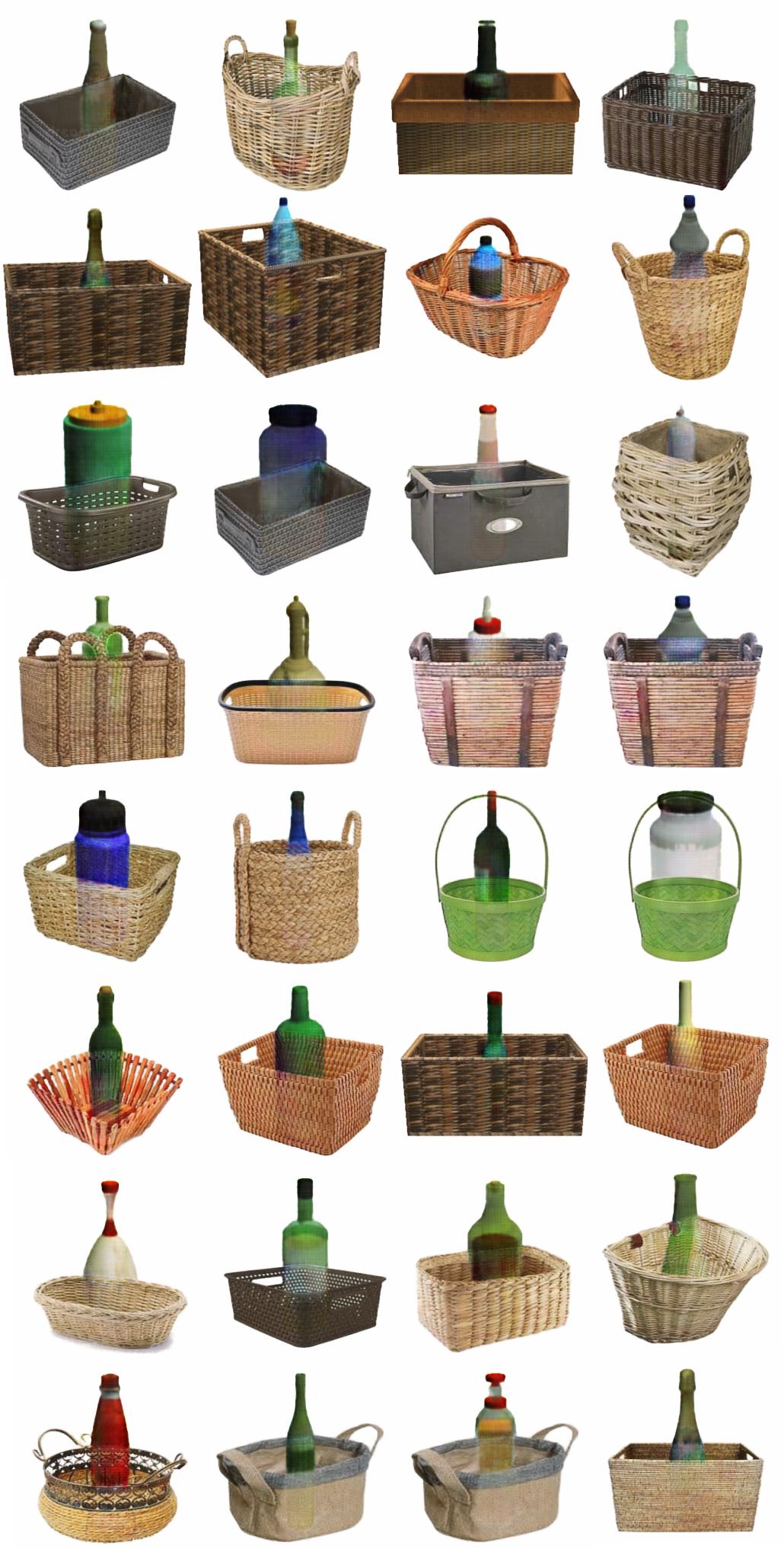}
  \caption{More results generated by the proposed network (DepGAN) on the bottle/basket shapenet\cite{shapenet} dataset.}
  \label{fig:resultsfullbottle}
\end{figure*}

\begin{figure*}[ht]
\centering
  \includegraphics[width=14cm]{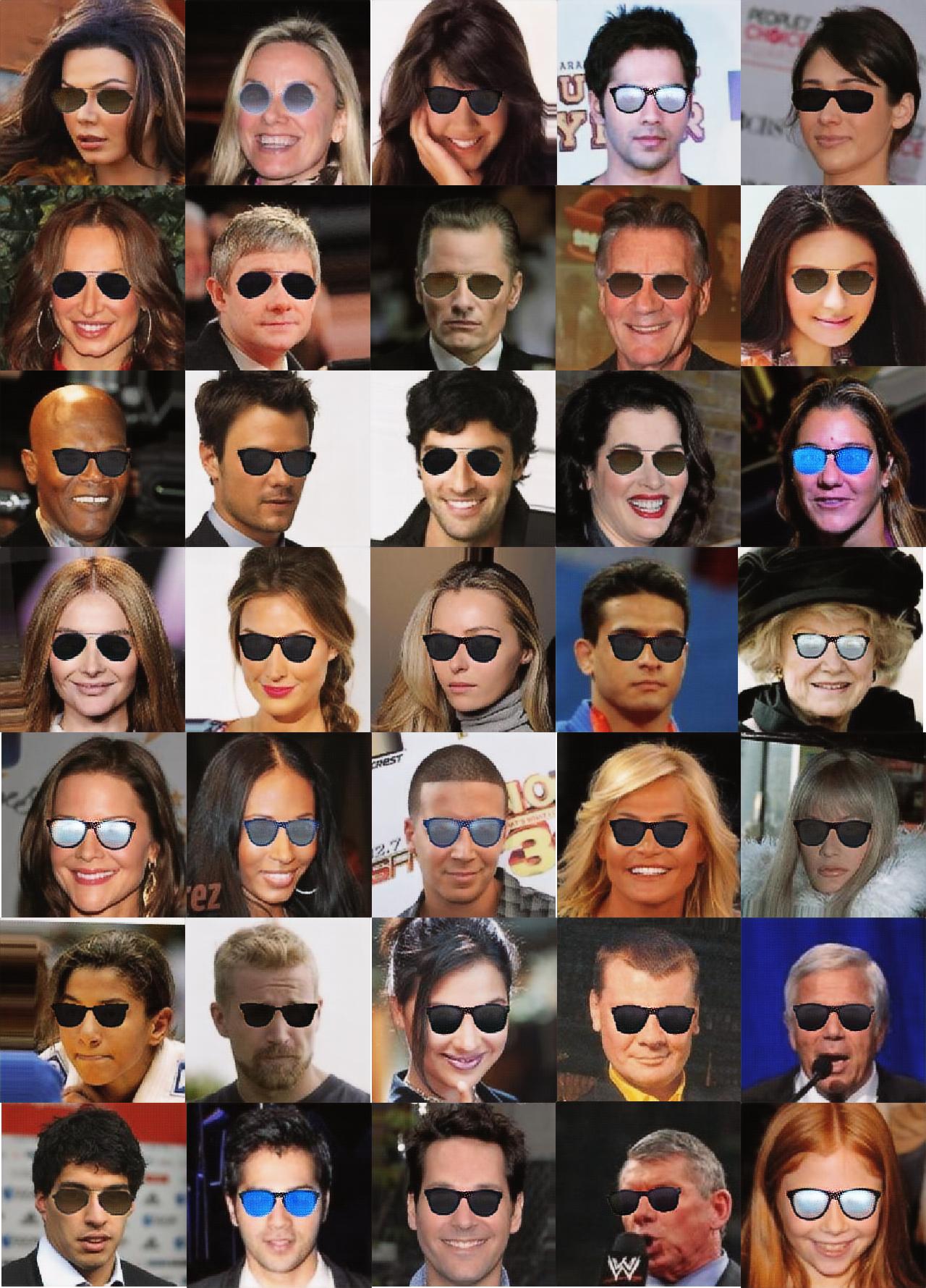}
  \caption{More results generated by the proposed network (DepGAN) on the face/glasses~\cite{STRAT} dataset.}
  \label{fig:resultsfullglasses}
\end{figure*}

\section{More Results}
\begin{table}[h]
\centering
\caption{The number of images in each dataset used in the evaluation.}
\begin{tabular}{lllll}
\hline
\textbf{Dataset}    & \textbf{Number of Images} \\ \hline
Shapenet (chair-desk)          & 1023 \\ \hline
Shapenet (bottle-basket)       & 121  \\ \hline
STRAT (face-glasses)           & 3000 \\ \hline
\end{tabular}
\label{tab:benchdataset}
\end{table}

\begin{table}[h]
\centering
\caption{The train/test/validation splits used for each dataset.}
\begin{tabular}{llll}
\hline
\textbf{Dataset}    & \textbf{Train} & \textbf{Test} & \textbf{Validation} \\ \hline
Shapenet (chair-desk)          & 80\%  & 10\%   & 10\% \\ \hline
Shapenet (bottle-basket)       & 70\%  & 15\%   & 15\%  \\ \hline
STRAT (face-glasses)           & 50\%  & 25\%   & 25\% \\ \hline
\end{tabular}
\label{tab:traintestsplit}
\end{table}

In this section, we provide additional image composition results on STRAT's \cite{STRAT}, and shapenet datasets generated by DepGAN. STRAT dataset provides diverse foreground images undergoing various transformations, allowing us to assess our network's capability to handle different transformation scenarios effectively. ShapeNet's dataset \cite{shapenet} offers synthetic datasets tailored specifically for occlusion testing. These datasets enabled us to assess our network's performance under varying degrees of occlusion complexity. Refer to Table \ref{tab:benchdataset} for the number of images used in each dataset and Table \ref{tab:traintestsplit} for the train/test/validation splits. Additional composition results on Shapenet and STRAT's datasets generated by the proposed DepGAN are presented in Figures \ref{fig:resultsfulldesk}-\ref{fig:resultsfullglasses}.

\begin{figure}[h]
  \includegraphics[width=\linewidth]{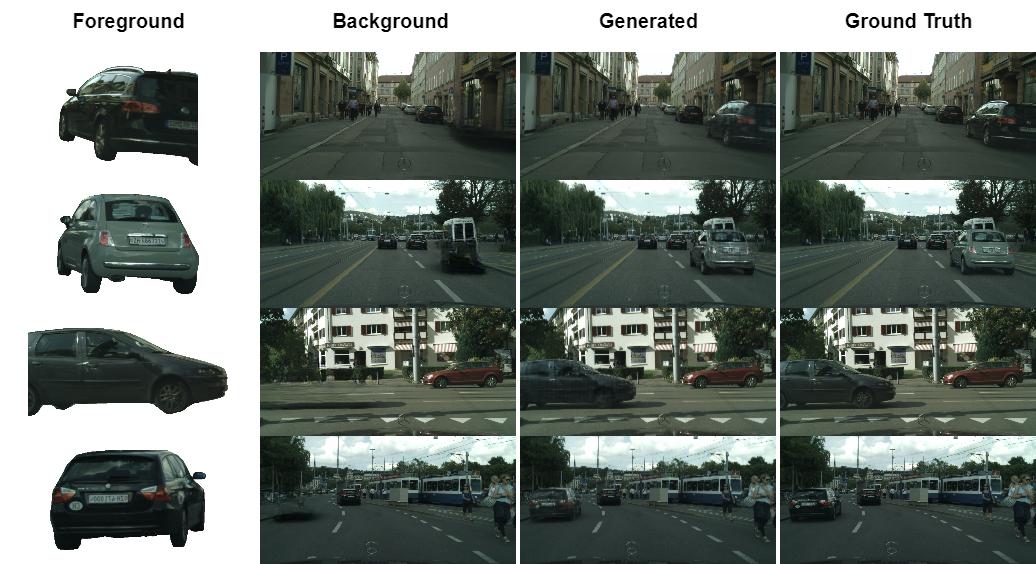}
  \caption{The composition results when trained with paired samples, where the foreground was removed, the whitespace inpainted, then placed back in.}
  \label{fig:realcar1}
\end{figure}
To further evaluate the effectiveness of DepGAN with transformations, we conducted an additional experiment using the car-street dataset. In this experiment, we aimed to test the model's capability to accurately scale and reintegrate a foreground object back into its original position within a sample image, ensuring that the result appears natural and consistent with the surrounding context. The car-street dataset consists of images featuring cars in various street settings. For this experiment, we created paired samples by removing the car (foreground object) from each image and then attempting to place it back using DepGAN. The primary goal was to assess whether DepGAN could accurately scale the car and position it in such a way that it seamlessly blended with the background, maintaining the correct perspective and proportions, results can be seen in Fig. \ref{fig:realcar1}.

\section{Network Architecture - Details}
\begin{figure}[ht]
  \includegraphics[width=\linewidth]{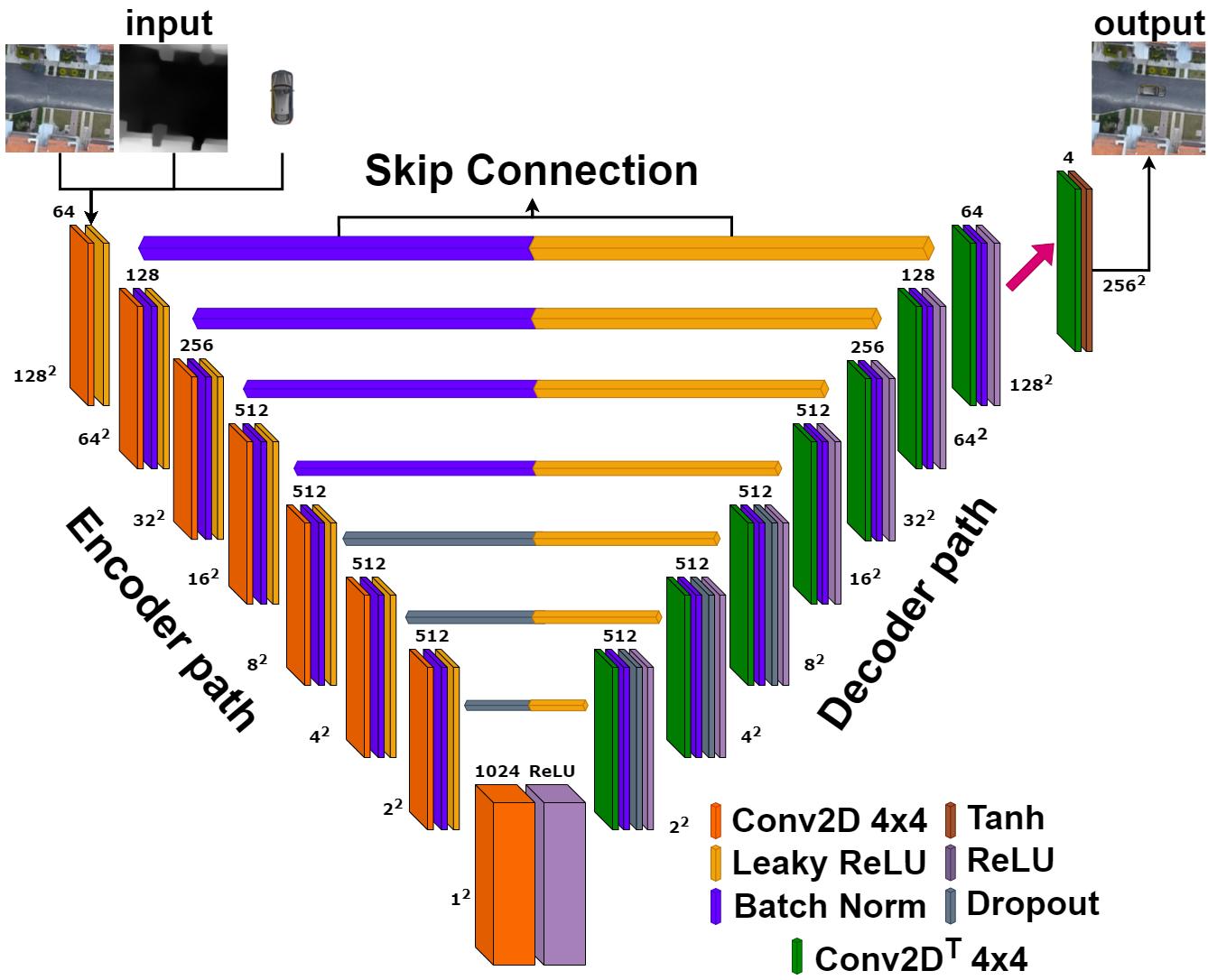}
  \caption{The architecture of the U-Net part of the generator. The Spatial Transformer
Network\cite{stn} is ran on the foreground input before the first encoder block of U-net.The Depth Aware Loss which penalizes the generator when generating foreground regions in composite image where the varying depth values are lighter is our additions to the network.}
  \label{fig:gen}
\end{figure}

\paragraph{Generator}
A detailed outline of the U-net part of our generator for ease of implementation is provided in Fig. \ref{fig:gen}.
\paragraph{Discriminator}
\begin{figure}[h]
\centering
\includegraphics[width=\linewidth]{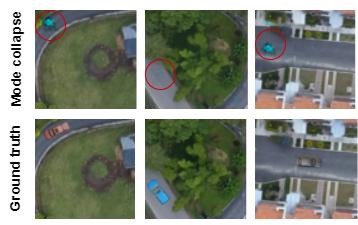}
\caption{In some datasets, the generator is able to fool the discriminator by producing a limited set of data patterns that do not resemble the ground truth, by making the discriminator stronger, we are able to combat this.}
\label{fig:modecollapse}\end{figure}
Initially, the discriminator architecture comprised layers with dimensions 64-128-256-512. However, an additional 512-block was introduced to the discriminator architecture to address scenarios where the generator outperformed the discriminator. In Fig. \ref{fig:modecollapse}, the generator deviates towards generating patterns aimed at deceiving the discriminator, while neglecting the goal of accurately replicating the ground truth images. This architectural adjustment resulted in more stable training dynamics, enhancing the overall performance and convergence of DepGAN, enabling the discriminator to better discern between ground truth images and generated images that merely mimic patterns to deceive it.

\begin{figure}[h]
\includegraphics[width=\linewidth]{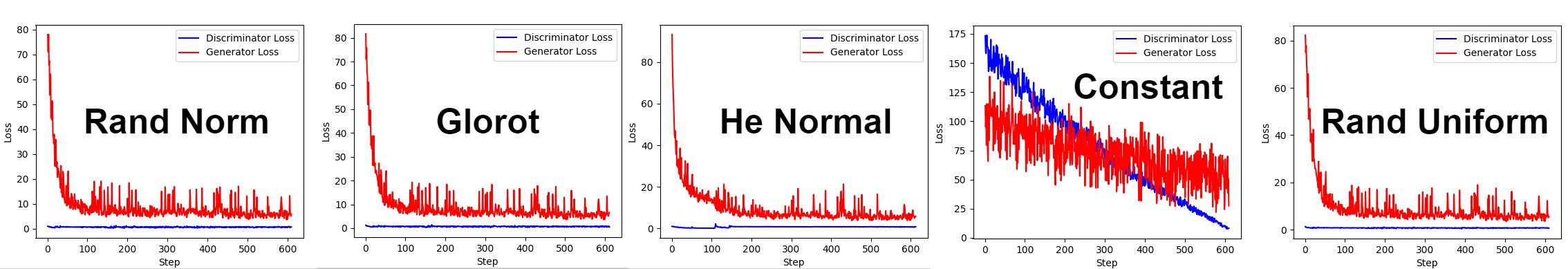}
\caption{Learning curves comparing the training loss initialized with different weight initializers. The model was trained on the same dataset using a fixed batch size, allowing for a comparison of convergence behavior and generalization performance.}
\label{fig:inits}\end{figure}

\paragraph{Weight Initialization}
In the pursuit of identifying the most effective weight initialization strategy, we explored various techniques. Five different weight initializers  such as Random Normal, Xavier Glorot, He, Uniform Distribution, Constant Initialization were tested. Among these initializers, Random Normal initialization with a standard deviation of 0.02 demonstrated the most promising results. This approach proved effective in achieving stable and consistent performance, particularly in accurately rendering background scenes. However, the other initialization methods encountered challenges in achieving comparable results, as shown in Fig. \ref{fig:inits}.

Uniform Distribution Initialization, Xavier Glorot Initialization and He Initialization, while commonly used in neural network initialization, did not yield satisfactory outcomes for DepGAN. These methods struggled to effectively capture the nuances of foreground-background integration, resulting in suboptimal performance and reduced fidelity in the composited images. Constant Initialization also faced limitations in effectively initializing the network weights. This approach failed to provide the necessary diversity and adaptability in weight initialization, leading to difficulties in learning complex features and spatial relationships. Random Normal initialization initializes weights from a normal distribution with a mean of 0 and a small standard deviation, typically around 0.02 in the case of DepGAN. Its success likely stems from the fact that this approach provides a degree of randomness to weight initialization, which helps prevent the network from getting stuck in local minima and promotes exploration of the weight space during training. 

\section{3D to 2D Composition - Application}
\begin{figure}[h]
  \includegraphics[width=\linewidth]{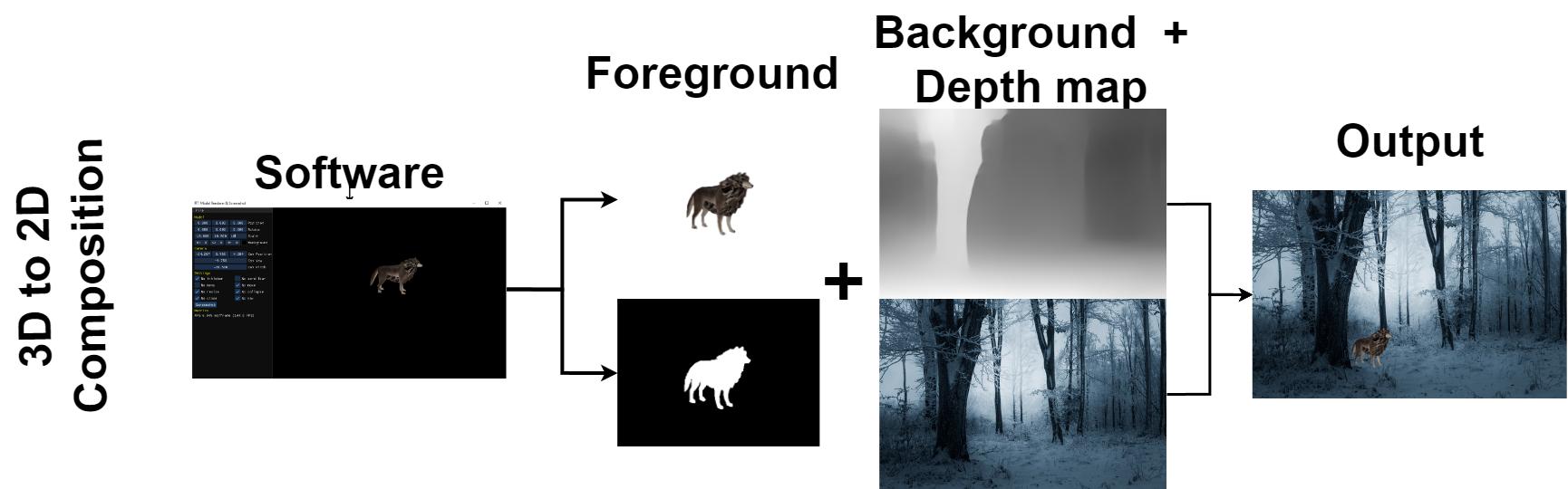}
  \caption{A visual representation outlining the sequence within our framework: initially, our proprietary program loads a 3D model and selects a particular viewpoint for integration into a foreground image. The foreground, combined with the background image and its associated depth map, is then inputted into DepGAN, resulting in the generation of the composite image.}
  \label{fig:framework}
\end{figure}

\begin{figure}[h]
  \includegraphics[width=\linewidth]{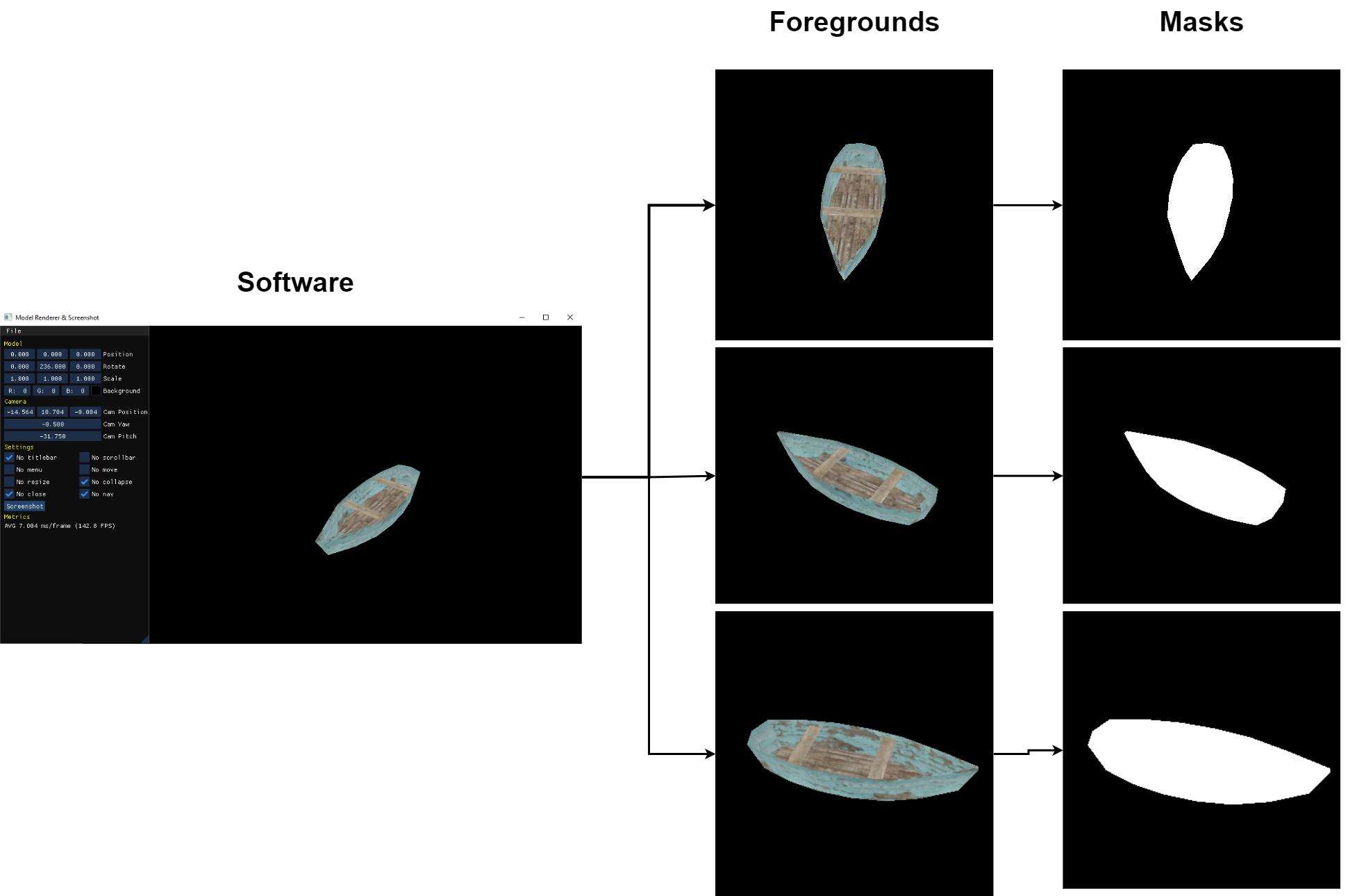}
  \caption{Our program to render a 3D model and capture a 2D image to be used as a foreground image in a dataset, capturing the model from a certain angle will generate a set of images, the foreground image captured from multiple views and their corresponding masks.}
  \label{fig:prog}
\end{figure}

\begin{figure}[h]
  \includegraphics[width=\linewidth]{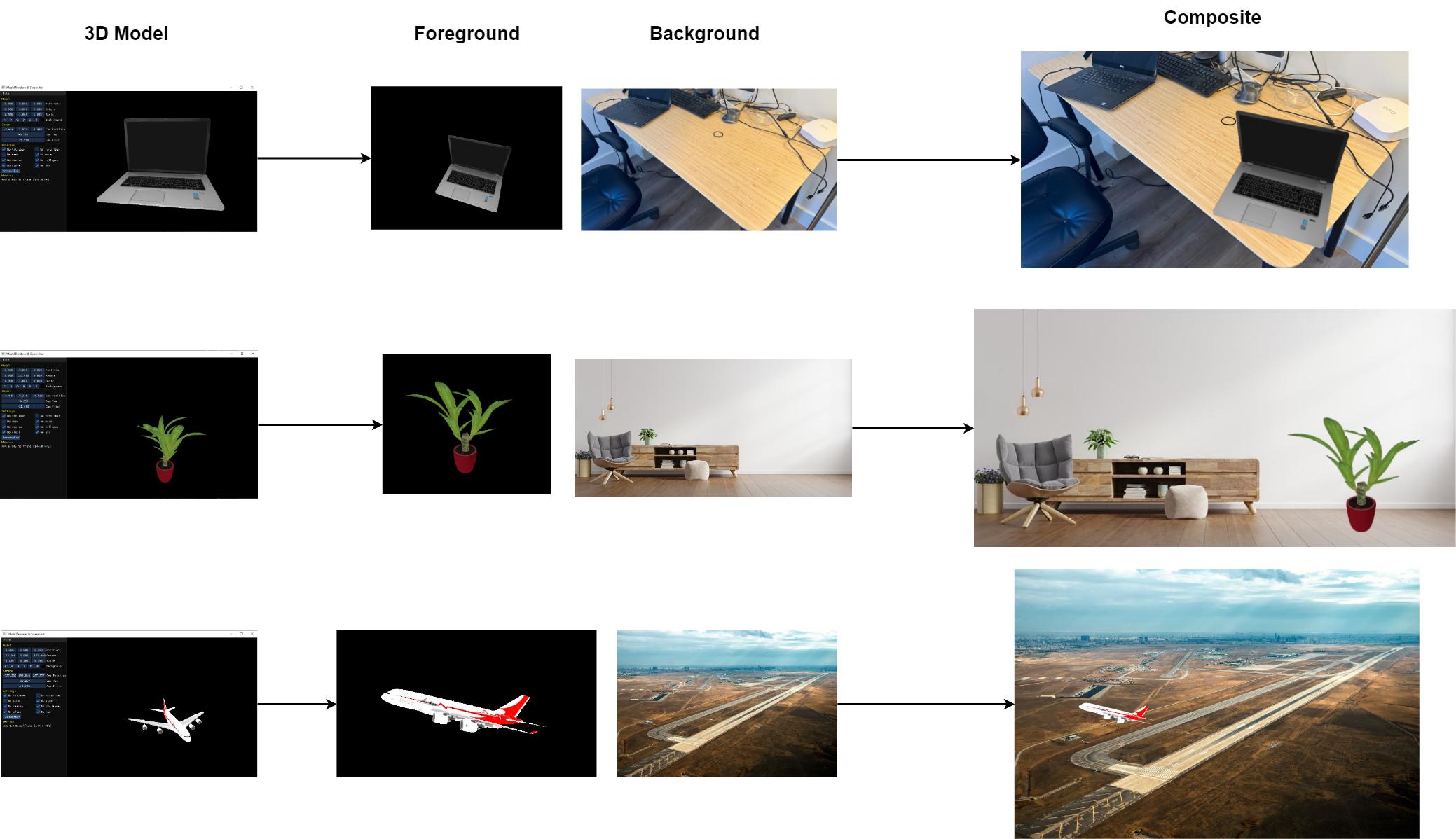}
  \caption{Some examples of compositing 3D models to 2D background images, first we render the model, capture a viewpoint which aligns with the background, then feed it along with the depth map and mask to generate the resulting composite images.}
  \label{fig:3dto2d}
\end{figure}

In this section, we demonstrate the utility of DepGAN for a potential application, i.e., compositing 3D models in 2D images. To be more specific, we capture the silhouettes of 3D models from specific viewpoints and composite with background images. To do that, we need a method to convert these models into 2D representations that align with the background image. So we develop a special software designed to render 3D models and capture them from multiple viewpoints, as shown in Fig. \ref{fig:prog}. When the viewpoints are captured, our software automatically generates masks for each viewpoint. We do this by setting the background color to a color that does not exist in the foreground image, then converting all pixels that do not match this color to white, and reverting the background color back to black. By automating the rendering and capture of 3D models and their corresponding masks, we significantly reduce the time and effort needed to create a dataset. To get the background images, we used the dataset by Ballesteros et al. \cite{roadds}, consisting of drone view imagery of roads, and to generate the depth maps, we employed MiDaS \cite{midas}. All the mentioned input, is then fed into DepGAN's generator, as shown in Fig. \ref{fig:framework}. Examples of 3D to 2D compositions using DepGAN can be seen in Fig. \ref{fig:3dto2d}. In these examples, we obtained shots of the objects from angles that seamlessly blend into the scene, such as capturing the laptop on the desk from a slightly elevated perspective. As a result of aligning the perspectives, the composited images appear authentic.